\documentclass[conference]{IEEEtran}
\IEEEoverridecommandlockouts
\usepackage{cite}
\usepackage{amsmath,amssymb,amsfonts}
\usepackage{algorithmic}
\usepackage{graphicx}
\usepackage{textcomp}
\usepackage{xcolor}

\usepackage{multirow}
\usepackage{url}

\usepackage[ruled,vlined,linesnumbered,noresetcount]{algorithm2e}
\SetKwInput{KwIn}{Input}
\SetKwInput{KwOut}{Output}

\usepackage{tikz}
\usetikzlibrary{arrows.meta,positioning,decorations.pathreplacing,calc}

\tikzset{
  box/.style={draw, rounded corners, line width=0.6pt, inner sep=4pt},
  mod/.style={box, minimum width=0.45\textwidth, align=center},
  thinarr/.style={-Latex, line width=0.7pt},
  dasharr/.style={-Latex, densely dashed, line width=0.7pt},
  lbl/.style={font=\footnotesize},
}

\def\BibTeX{{\rm B\kern-.05em{\sc i\kern-.025em b}\kern-.08em
    T\kern-.1667em\lower.7ex\hbox{E}\kern-.125emX}}
\begin{document}

\title{Heterogeneous Multi-Agent \\Proximal Policy Optimization
for \\ Power Distribution System Restoration\\
{\footnotesize \textsuperscript{}}
\thanks{This research is supported by the National Science Foundation (NSF) under grant ECCS-2223628.}
}

\author{
\IEEEauthorblockN{1\textsuperscript{st} Parya Dolatyabi}
\IEEEauthorblockA{\textit{Department of Computer Science} \\
\textit{University of Tulsa}\\
Tulsa, USA \\
parya-dolatyabi@utulsa.edu}
\and
\IEEEauthorblockN{2\textsuperscript{nd} Ali Farajzadeh Bavil}
\IEEEauthorblockA{\textit{Department of Computer Science} \\
\textit{University of Tulsa}\\
Tulsa, USA \\
ali-farajzadeh@utulsa.edu}
\and
\IEEEauthorblockN{3\textsuperscript{rd} Mahdi Khodayar}
\IEEEauthorblockA{\textit{Department of Computer Science} \\
\textit{University of Tulsa}\\
Tulsa, USA \\
mahdi-khodayar@utulsa.edu}
}

\maketitle

\begin{abstract}
Restoring power distribution systems (PDSs) after large-scale outages requires sequential switching actions that reconfigure feeder topology and coordinate distributed energy resources (DERs) under nonlinear constraints, including power balance, voltage limits, and thermal ratings. These challenges limit the scalability of conventional optimization and value-based reinforcement learning (RL) approaches. This paper applies a Heterogeneous-Agent Reinforcement Learning (HARL) framework via Heterogeneous-Agent Proximal Policy Optimization (HAPPO) to enable coordinated restoration across interconnected microgrids. Each agent controls a distinct microgrid with different loads, DER capacities, and switch counts. Decentralized actors are trained with a centralized critic for stable on-policy learning, while a physics-informed OpenDSS environment enforces electrical feasibility. Experiments on IEEE 123-bus and 8500-node feeders show HAPPO outperforms PPO, QMIX, Mean-Field RL, and other baselines in restored power, convergence stability, and multi-seed reproducibility. Under a 2400 kW generation cap, the framework restores over 95\% of available load on both systems with low-latency execution, supporting practical real-time PDS restoration.
\end{abstract}


\section{Introduction}
Modern power distribution systems face increasing exposure to environmental hazards, component aging, and cybersecurity threats, resulting in frequent service interruptions \cite{casey}. Distributed energy resources (DERs) such as photovoltaics, microturbines, and storage can accelerate local recovery \cite{rhodes}, yet their limited capacity and the voltage, thermal, and topological constraints of distribution feeders complicate safe and efficient restoration \cite{poudel}. Recent advancements in machine learning for critical infrastructure highlight the growing role of data-driven recovery, including transformer–reinforcement learning frameworks for large-scale repair sequencing \cite{amani} and ensemble techniques for high-fidelity ground-motion prediction \cite{seismic}.

Conventional restoration is typically formulated as a mixed-integer nonlinear optimization problem \cite{pham,song,aravena}, which may provide optimal switching plans but becomes computationally prohibitive for large or dynamically changing feeders \cite{liao}. Moreover, restoration is inherently sequential and stochastic, with each switching action altering future feasibility \cite{shi}. Reinforcement learning (RL) provides an alternative by learning control policies directly through interaction \cite{zhang}, supported by broader advances in deep and graph-based learning for power systems \cite{powerGNNreview}. Early RL efforts used single-agent controllers based on DQN \cite{igder} or actor–critic methods \cite{yang}, but suffered from biased value estimates \cite{yu}, inefficient exploration, and limited scalability \cite{selim}.

Multi-agent reinforcement learning (MARL) addresses scalability by assigning local controllers to network subregions. Methods such as MAES \cite{zhang2}, MAGDPG \cite{fan}, Mean-Field RL \cite{zhao}, and QMIX \cite{si} improve coordination but face key limitations: (i) reliance on value-based updates or fixed exploration rules, amplifying approximation errors; (ii) feasibility enforcement via action masking or early termination, disrupting learning; and (iii) overfitting to common operating patterns.

To overcome these challenges, this paper applies Heterogeneous-Agent Proximal Policy Optimization (HAPPO) within a heterogeneous-agent reinforcement learning (HARL) architecture \cite{HARL_JMLR} for power distribution system restoration. HAPPO extends PPO to multi-agent settings through sequential per-agent updates guided by a centralized critic, enabling stable learning among microgrids with differing load levels, generation capacities, and switch counts. These structural differences make agents non-exchangeable and unsuitable for parameter-sharing methods such as MAPPO.

We develop an OpenDSS-based simulation environment \cite{opendss} performing full three-phase power-flow analysis after each switching action. Rather than masking infeasible actions \cite{vu} or terminating episodes \cite{yu}, the environment applies differentiable penalty terms proportional to voltage, current, and generation-limit violations, while enforcing a system-level generation cap and local microgrid balance constraints. This soft-constraint design provides dense feedback and allows agents to learn corrective behavior.

Experiments on IEEE 123-bus \cite{123} and 8500-node \cite{8500} feeders show the proposed HAPPO framework achieves faster convergence, higher restored power, and improved stability compared with DQN \cite{igder}, PPO \cite{zhou}, MAES \cite{zhang2}, MAGDPG \cite{fan}, MADQN \cite{vu}, Mean-Field RL \cite{zhao}, and QMIX \cite{si}. The contributions are: (1) first application of HAPPO to feeder-level restoration with structurally heterogeneous microgrid agents; (2) a stable and scalable learning framework combining decentralized actors with a centralized advantage critic.

\section{Problem Formulation}
\label{sec:problem_formulation}
\subsection{Network Model and Decision Variables}
Let the distribution network be represented by a graph 
$\mathcal{G} = (\mathcal{N}, \mathcal{E})$, where $\mathcal{N}$ denotes the set of buses and $\mathcal{E}$ denotes the set of branches. Loads lie on $\mathcal{D} \subseteq \mathcal{N}$ and distributed energy resources (DERs) on 
$\mathcal{G} \subseteq \mathcal{N}$. A subset 
$\mathcal{S} \subseteq \mathcal{E}$ contains controllable sectionalizing or tie 
switches. Each switch $q \in \mathcal{S}$ has a binary state $b_q \in \{0,1\}$ indicating a 
\emph{closed} (energized) or \emph{open} (de-energized) line. After an outage, 
restoration proceeds through discrete switching actions that reconfigure the 
topology and determine the energized buses and restored loads.

\subsection{Scenario Randomization and Priorities}
Each simulation episode introduces variability by randomizing the faulted or unavailable branches, the placement of critical loads, and any optional changes in DER penetration levels. Each load $k \in \mathcal{D}$ is assigned a priority weight
$c_k \in \{1,\dots,10\}$, indicating its relative importance in the restoration
process.  
These priority assignments differ between feeders and determine the 
weighted-load objective defined below.

\subsection{Restoration Objective and Operational Constraints}

Let $P_k$ denote the restored active power at load bus $k$.  
The system-level restoration objective is to maximize the 
weighted restored power:
\begin{equation}
    \max_{\{b_q\}} 
    J = \sum_{k \in \mathcal{D}} c_k P_k ,
    \label{eq:objective}
\end{equation}
where $J$ measures the total service recovery prioritized by load criticality.

At each switching configuration, the distribution system must satisfy
fundamental electrical and operational limits. These constraints define the feasible space of valid switching configurations
during restoration.

\subsubsection{Power Balance}
The total restored load and network losses must not exceed the available DER
generation. Here, the inequality permits unused generation capacity during partial restoration:
\begin{equation}
\sum_{k \in \mathcal{D}} P_k 
+ \sum_{e \in \mathcal{E}} P^{\text{loss}}_e
\le \sum_{g \in \mathcal{G}} P_g ,
\label{eq:power_balance}
\end{equation}
where $P_k$ is the active power consumed by load $k$, $P^{\text{loss}}_e$ is the power loss on branch $e$, and $P_g$ is the active power generated by DER unit $g$.

\subsubsection{Voltage Limits}
Voltage magnitude at each bus must remain within allowable bounds:
\begin{equation}
V_{\min} \le V_k \le V_{\max}, 
\quad \forall k \in \mathcal{N},
\label{eq:voltage_limits}
\end{equation}
where $V_k$ is the voltage magnitude at bus $k$.

\subsubsection{DER Operating Limits}
Each DER must operate within its active and reactive power bounds:
\begin{equation}
\scalebox{0.85}{$
P_g^{\min} \le P_g \le P_g^{\max}, \quad 
Q_g^{\min} \le Q_g \le Q_g^{\max}, 
\quad \forall g \in \mathcal{G},
\label{eq:der_limits}$}
\end{equation}
where $P_g$ and $Q_g$ are the active and reactive power outputs of DER unit $g$, respectively.

\subsubsection{Thermal Line Limits}
Branch power flows must satisfy thermal constraints:
\begin{equation}
P_e^2 + Q_e^2 \le (S^{\max}_e)^2, 
\quad \forall e \in \mathcal{E},
\label{eq:thermal_limits}
\end{equation}
where $P_e$ and $Q_e$ are the active and reactive power flows on branch $e$, and $S^{\max}_e$ is the thermal capacity of branch $e$.

\subsubsection{Global Generation Capacity}
The total DER output across all microgrids cannot exceed the system-wide
generation cap:
\begin{equation}
\sum_{g \in \mathcal{G}} P_g \le 2400~\text{kW}.
\label{eq:global_gen_cap}
\end{equation}

\subsubsection{Local Microgrid Balance}
For each microgrid $i$, local load consumption cannot exceed its local DER
generation:
\begin{equation}
P^{\text{load}}_i \le P^{\text{gen}}_i, 
\quad \forall i.
\label{eq:local_balance}
\end{equation}

\subsection{Structural Heterogeneity and Feeder Partitioning}

To reduce the combinatorial complexity of the restoration problem, the feeder is partitioned into $N$ microgrids. Each microgrid $i$ includes its own subset of loads and DERs, operates a distinct set of controllable switches $\mathcal{Q}_i \subseteq \mathcal{S}$, and exhibits unique load magnitudes and generation capacities, resulting in structurally heterogeneous operating regions. While each agent controls switches with identical action types (open/close/no-op), microgrids differ substantially in the number of switches they control, their load profiles, DER availability, and topological structure. This structural heterogeneity makes microgrids non-exchangeable and motivates the heterogeneous-agent formulation adopted in later sections.

Microgrids are defined following standard feeder partitioning practices based on topology and DER locality, consistent with prior restoration benchmarks to ensure fair comparison. This choice reflects physical connectivity and operational coupling, and alternative reasonable partitions yielded similar convergence behavior, indicating low sensitivity to minor variations.

\subsection{Reward Function}

When solving~\eqref{eq:objective} via reinforcement learning, a dense
reward function is constructed to encourage restoration progress while penalizing
constraint violations. Let $\Delta P_t^{\text{rest}} = \sum_{k \in \mathcal{D}} c_k (P_{k,t} - P_{k,t-1})$ denote the weighted incremental increase in restored power at step $t$, where $P_{k,t}$ is the power restored to load $k$ at time $t$. Then $r_t = 
\alpha\, \Delta P^{\text{rest}}_t
- \beta \frac{P^{\text{loss}}_t}{P_{\text{gen}}}
- \lambda\, \xi_t ,$ where $\Delta P^{\text{rest}}_t$ represents the incremental increase in restored power, and $P^{\text{loss}}$ is normalized by the fixed generation cap 
$P_{\text{gen}} = 2400~\text{kW}$. The penalty term $\xi_t$ collects the magnitudes of violations of constraints~\eqref{eq:power_balance}--\eqref{eq:local_balance}, whereas the parameters $\alpha$, $\beta$, and $\lambda$ serve as nonnegative weights that tune the contributions of restoration progress, loss minimization, and constraint enforcement, respectively. This reward formulation preserves the mathematical objective while enabling 
learning-based optimization in Section~\ref{sec:methodology}.\section{Proposed HAPPO-Based Methodology}
\label{sec:methodology}
\subsection{HARL Architecture Overview}
The proposed approach is built within a heterogeneous-agent reinforcement learning (HARL) architecture \cite{HARL_JMLR}, where each microgrid agent maintains an independent actor policy and a shared centralized critic estimates global value functions. This design reflects the physical hierarchy of distribution systems: local agents observe and control only their own microgrids, while the critic leverages system-wide information to provide coordinated learning signals. Each agent~$i$ follows a policy $a_{i,t} \sim \pi_{\theta_i}(o_{i,t}), $ issuing a discrete switching action (\textit{open}, \textit{close}, or \textit{no-op}). The centralized critic $V_\phi(s_t)$ processes the global system state to compute value and advantage estimates that guide all agents.

A defining feature of HAPPO is its sequential per-agent update rule: agents update their policies one at a time, with each update conditioned on the latest policies of previously updated agents. This reduces gradient interference and enhances stability in strongly coupled multi-agent settings such as distribution feeders.

\begin{figure}[t]
    \centering
    \includegraphics[width=0.48\textwidth]{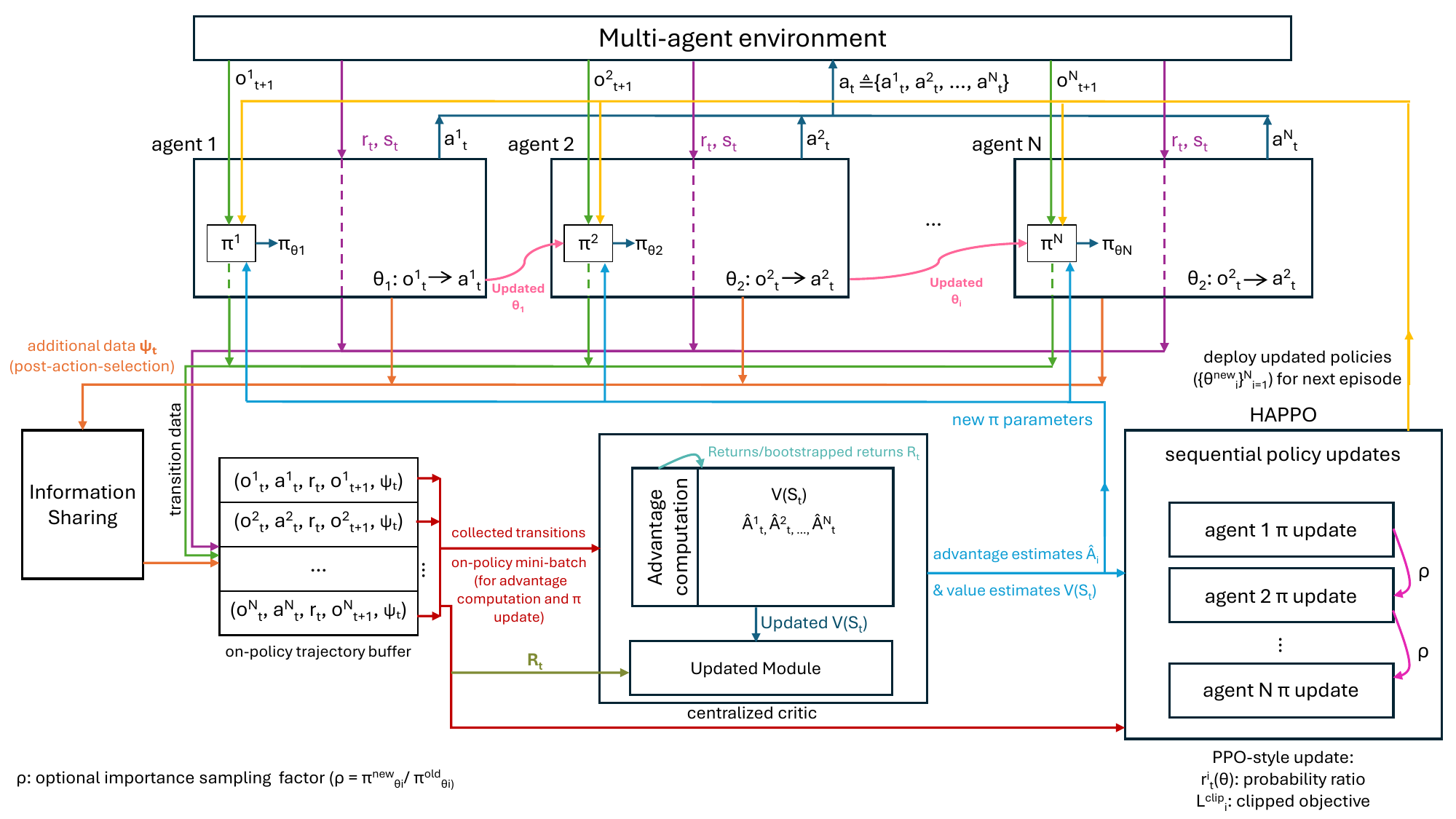}
    \caption{Overview of the proposed HAPPO-based HARL framework for multi-agent power distribution system restoration. 
    Individual microgrid actors issue switching actions based on local observations, 
    while a shared centralized critic computes value and advantage estimates from global states. 
    OpenDSS \cite{opendss} simulation feedback enables constraint-aware sequential policy updates.}
    \label{fig:happo_architecture}
\end{figure}

\subsection{Agent--Environment Interaction}

At each simulation step~$t$, each agent $i$ receives a local observation $o_{i,t}$ 
(e.g., voltages, currents, DER outputs, and switch states), 
selects an action $a_{i,t} \sim \pi_{\theta_i}(o_{i,t})$, 
and the joint action $a_t = (a^1_t, \ldots, a^N_t)$ is executed in the OpenDSS simulator. OpenDSS computes the resulting feeder state, including next global state $s_{t+1}$, updated local observations $\{o_{i,t+1}\}$, restored power and system losses, voltage and current magnitudes, and the shared global reward $r_t$.

Each transition contributes to an on-policy trajectory buffer $(o^i_t, a^i_t, r_t, o^i_{t+1}, \psi_t),$ where $\psi_t$ contains auxiliary information such as the joint action $\mathbf{a}_t = (a_{1,t}, \ldots, a_{N,t})$ and other global signals used for centralized training.  
These trajectories support centralized advantage computation and sequential policy optimization.

\subsection{Advantage Computation via $\lambda$-GAE}

To quantify each agent's temporal contribution to the global return, 
the framework employs the generalized advantage estimator (GAE):
\begin{align}
\delta_t^i &= r_t + \gamma V_\phi(s_{t+1}) - V_\phi(s_t), \\
\hat{A}^i_t &= \sum_{\ell=0}^{T-t-1} (\gamma\lambda)^\ell \, \delta_{t+\ell}^i,
\end{align}
where $\gamma$ is the discount factor and $\lambda\in[0,1]$ controls the bias--variance trade-off, and $\ell$ is the time-step offset. Note that while the TD error uses the global reward $r_t$ and centralized value function $V_\phi(s_t)$, advantages are computed separately for each agent $i$ to support sequential policy updates.
The centralized critic provides low-variance, system-consistent value estimates, enabling each agent to compute 
advantages aligned with the global restoration objective.

\subsection{Sequential Policy Updates (HAPPO)}

HAPPO performs \textit{sequential PPO-style policy updates} for agents $1$ through $N$.  
For agent~$i$, the clipped surrogate objective is:
\begin{equation}
\scalebox{0.75}{$
L_i(\theta_i) =
\mathbb{E}\!\left[
\min\!\left(
\rho_{i,t}\hat{A}^i_t,\,
\text{clip}(r_{i,t},1-\epsilon,1+\epsilon)\hat{A}^i_t
\right)
\right]
- \beta_{\mathrm{ent}}\,\mathbb{E}[H(\pi_{\theta_i})],
$}
\end{equation}

\noindent 
where the probability ratio is $\rho_{i,t} = \dfrac{\pi^{\text{new}}_{\theta_i}(a_{i,t}\mid o_{i,t})}
{\pi^{\text{old}}_{\theta_i}(a_{i,t}\mid o_{i,t})}.$ The entropy regularization term $H(\pi_{\theta_i})$ encourages exploration,
with $\beta_{\mathrm{ent}}$ controlling the strength of the entropy bonus.

The clipping term enforces a trust-region constraint, preventing large, destabilizing policy shifts, 
while the entropy term $\beta_{\mathrm{ent}}$ encourages adequate exploration.  
Sequential updates reduce gradient interference and support monotonic policy improvement. The critic parameters $\phi$ are trained by regressing $V_\phi(s_t)$ onto bootstrapped returns:
\begin{equation}
\scalebox{0.75}{$
L_{\text{critic}} = \mathbb{E}\left[(V_\phi(s_t) - \hat{R}_t)^2\right], \quad \hat{R}_t = \sum_{\ell=0}^{T-t-1} \gamma^\ell r_{t+\ell} + \gamma^{T-t} V_\phi(s_T).
$}
\end{equation}

\subsection{Training Workflow}

Algorithm~\ref{alg:happo_train_unified} summarizes the unified HAPPO training procedure for both IEEE~123-bus and 8500-node feeders. Each iteration involves: agents interact with OpenDSS for $T$ steps to generate rollouts; the centralized critic computes temporal-difference errors and $\lambda$-GAE advantages; agents sequentially update actor parameters via clipped PPO-style optimization; the critic regresses against bootstrapped returns to refine value estimates; and updated networks deploy for the next episode. This workflow enables seamless scaling between medium- and large-scale feeders, ensuring stable, reproducible learning dynamics.

\begin{algorithm}[t]
\footnotesize
\caption{Unified HAPPO Training Workflow for IEEE-123 and IEEE-8500 Feeders}
\label{alg:happo_train_unified}
\SetNlSty{textbf}{}{.}
\SetAlgoNlRelativeSize{-1}
\KwIn{Environment $\mathcal{E} \in \{\text{IEEE-123}, \text{IEEE-8500}\}$, configuration $\mathcal{C}$}
\KwOut{Trained actor and critic networks for all agents}
\BlankLine
\textbf{Initialize:} agents, critic $V_\phi$, rollout buffer $\mathcal{B}$, and logger\;
Load hyperparameters: rollout length $T$, learning rates, clipping threshold $\epsilon$, 
discount $\gamma$, GAE parameter $\lambda$, etc., and checkpoint interval $S$ (\textit{save period})\;
\BlankLine
\For{each iteration $u = 1$ \KwTo $U_{\max}$}{
    Reset environment and obtain $(o_{i,0}, s_0)$\;
    \For{$t = 0$ \KwTo $T-1$}{
        \ForEach{agent $i$}{
            Sample action $a_{i,t} \sim \pi_{\theta_i}(a|o_{i,t})$\;
        }
        Execute joint action $\mathbf{a}_t$ in OpenDSS\;
        Observe $(o_{i,t+1}, s_{t+1}, r_t, d_t, \psi_t)$\;
        Store transition in buffer $\mathcal{B}$\;
    }
    Compute TD errors and GAE advantages for each agent:
        $\delta_t^i = r_t + \gamma V_\phi(s_{t+1}) - V_\phi(s_t),
    \quad
    \hat{A}_t^i = \sum_{\ell=0}^{T-t-1} (\gamma\lambda)^\ell \delta_{t+\ell}^i$\;

    \BlankLine
    \ForEach{agent $i$ (sequential update)}{
        Freeze $\pi_{\theta_j}$ for all $j \neq i$\;
        \For{each PPO epoch}{
            Sample minibatches $(o_i, a_i, \hat{A}, \log\pi_{\text{old}})$\;
            Compute probability ratio $\rho$ and clipped surrogate objective\;
            Update actor parameters $\theta_i \leftarrow \theta_i - \alpha_\pi \nabla L_{\text{actor}}$\;
        }
    }
    Compute critic loss:
        $L_{\text{critic}} = \mathbb{E}\!\left[(V_\phi(s_t) - \hat{R}_t)^2\right]$,\\
    where $\hat{R}_t = \sum_{\ell=0}^{T-t-1} \gamma^\ell r_{t+\ell} + \gamma^{T-t} V_\phi(s_T)$;\;
    Update critic: $\phi \leftarrow \phi - \alpha_V \nabla L_{\text{critic}}$\;

    \BlankLine
    Log training metrics at iteration $u$ (\textit{iter})\;
    \If{$u \bmod S = 0$}{
        Save model checkpoint (\textit{ckpt})\;
    }
}
\Return trained $\{\pi_{\theta_i}\}_{i=1}^N$ and critic $V_\phi$\;
\end{algorithm}

\section{Experimental Setup}
\label{sec:exp_setup}

\subsection{Simulation Environment}
Experiments are conducted on the IEEE\,123-bus~\cite{123} and IEEE\,8500-node~\cite{8500} feeders using a custom Python--OpenDSS~\cite{opendss} environment that performs full three-phase AC power-flow calculations after each switching action. The simulator returns voltages, currents, DER outputs, losses, and restored load. Operational constraints, including voltage limits, thermal ratings, DER bounds, and the global generation cap \(P_{\text{gen}} = 2400\,\text{kW}\), are enforced through differentiable penalty terms rather than action masking or early termination, providing dense feedback near feasibility boundaries.

\begin{figure}[t]
    \centering
    \includegraphics[width=0.48\textwidth]{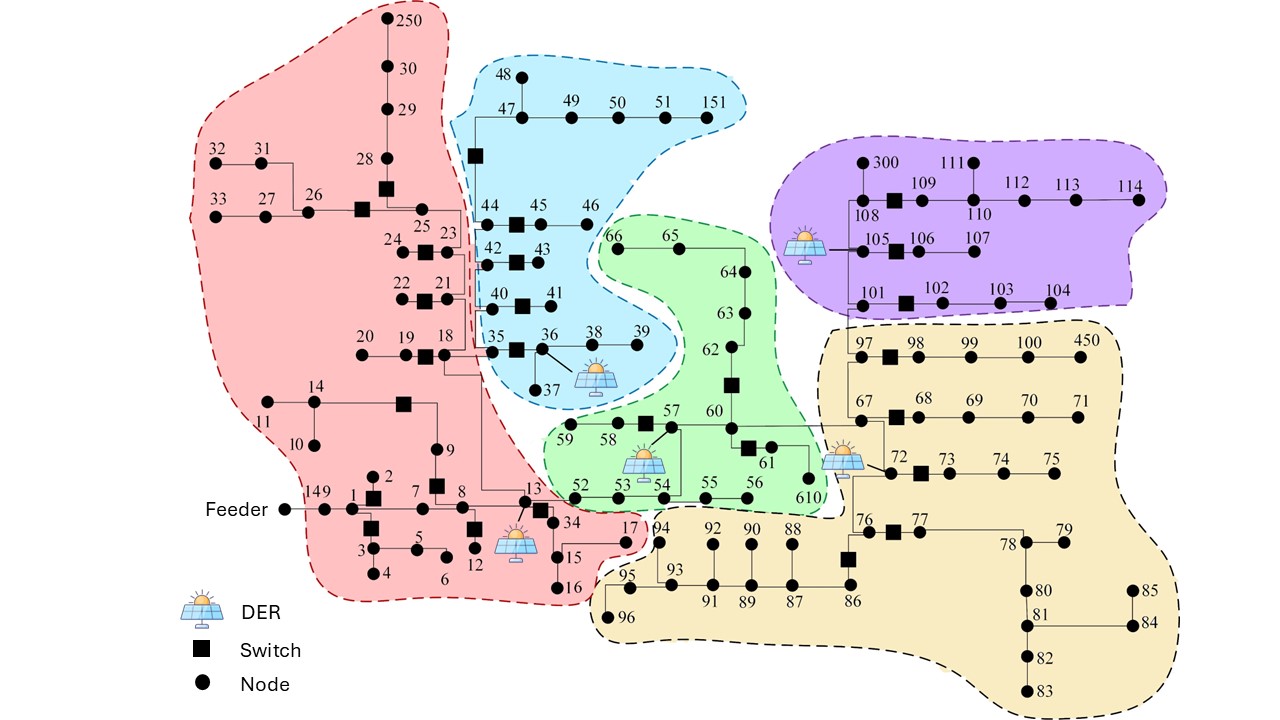}
    \caption{IEEE\,123-bus distribution feeder used in the experiments. 
    The system includes 26 controllable switches and five DERs 
    (total generation 2400\,kW), partitioned into five microgrid regions.}
    \label{fig:ieee123}
\end{figure}

The IEEE 123-bus feeder (Fig.~\ref{fig:ieee123}) has 123 buses, a 3025 kW load, five DERs totaling 2400 kW, and 26 controllable switches, partitioned into five agent-controlled microgrids. The IEEE 8500-node feeder contains about 8500 nodes, a 25 MW load, 100 switches, and 2400 kW of DER capacity, divided into ten microgrids. Its scale and deep radial structure make it a demanding testbed for HAPPO scalability.

\subsection{Training Configuration}
All experiments follow the unified HARL workflow in Section~\ref{sec:methodology} (Algorithm~\ref{alg:happo_train_unified}). Each $T$-step episode uses decentralized recurrent actors for discrete switching actions and a centralized critic \(V_\phi(s_t)\) for values and $\lambda$-GAE advantages. Policy learning employs PPO-style updates with clipping threshold $\epsilon$, entropy coefficient $\beta_{\text{ent}}$, and sequential per-agent optimization. Rewards combine incremental restored power, normalized loss penalties, and continuous constraint violations. Both feeders use a fixed generation cap \(P_{\text{gen}} = 2400\,\text{kW}\). Hyperparameters ($\gamma$, $\lambda$, learning rates, batch sizes) remain identical unless scale adjustments are needed.

\subsection{Hardware and Evaluation Metrics}
Training is performed on a workstation with an NVIDIA RTX\,6000 Ada GPU (48\,GB), AMD Threadripper PRO CPU, and 256\,GB RAM, with OpenDSS on CPU and PyTorch on GPU. Performance is evaluated using restored active power, convergence stability across seeds, inference latency, and training time, enabling consistent comparison across feeders and baseline RL/MARL methods.

\section{Numerical Results}
\label{sec:results}
\subsection{Performance on IEEE 123- and 8500-Bus Systems}
\begin{figure}[t]
    \centering
    \includegraphics[width=0.41\textwidth]{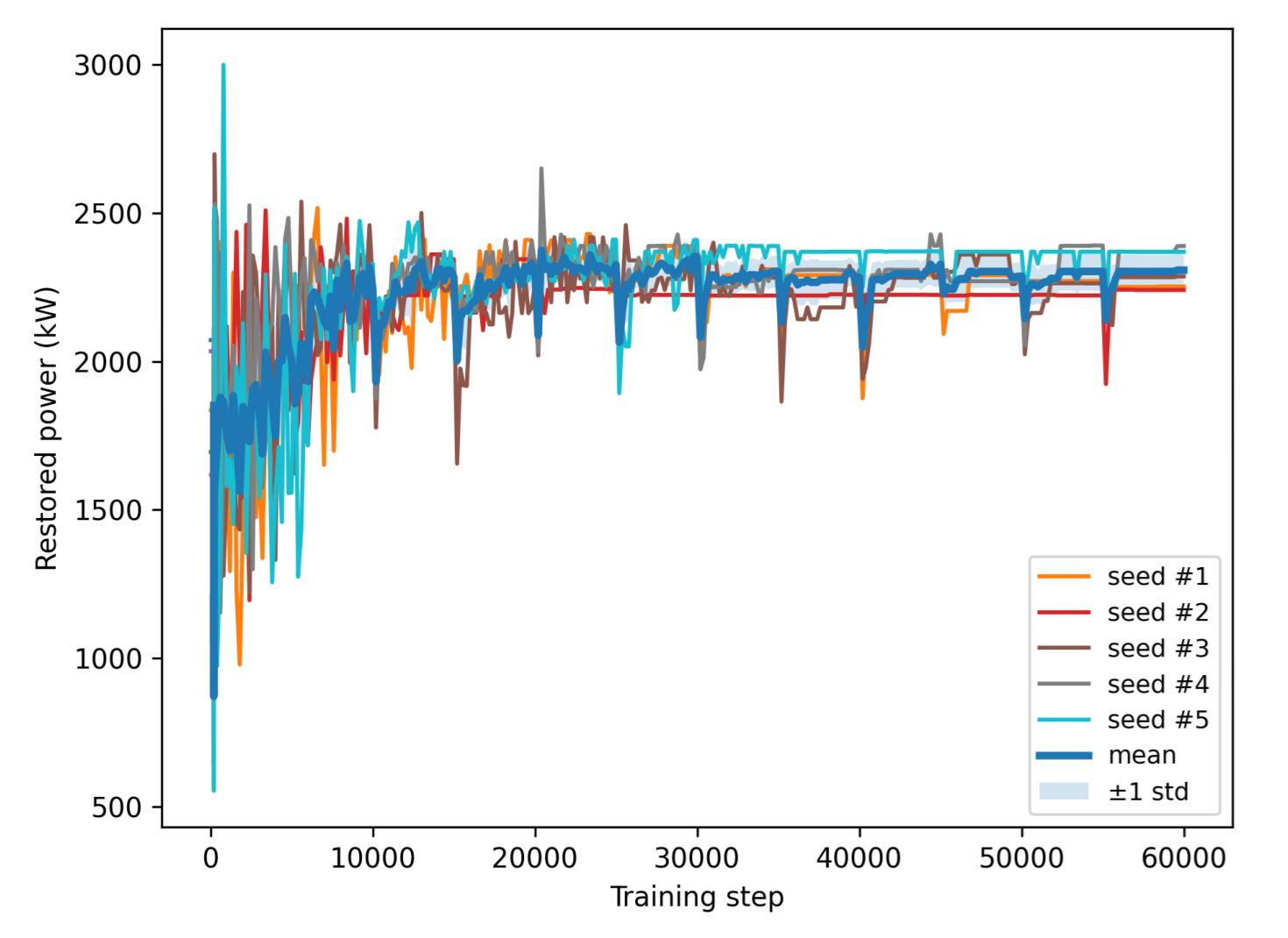}
    \caption{Training performance of HAPPO on the IEEE 123-bus feeder across five random seeds showing restored active power evolution.}
    \label{fig:results_123}
\end{figure}

\begin{figure}[t]
    \centering
    \includegraphics[width=0.41\textwidth]{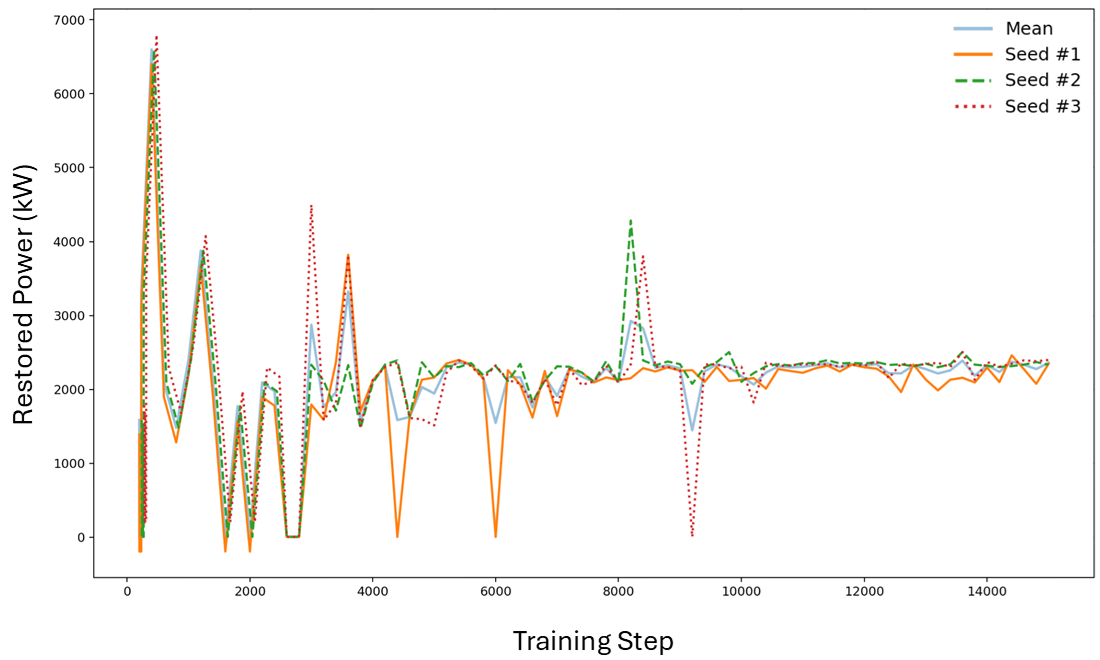}
    \caption{Training performance of HAPPO on the IEEE 8500-bus system across three random seeds showing restored active power evolution.}
    \label{fig:results_8500}
\end{figure}

Figures~\ref{fig:results_123} and~\ref{fig:results_8500} show the evolution of restored active power during training for the two feeders. Each curve corresponds to an independent seed initialized with randomized outage locations, critical-load placements, and DER conditions.

In the IEEE\,123-bus feeder, 5 agents coordinate 26 switches.
Across five independent random seeds, the restored power stabilizes at $2294 \pm 53~\text{kW}(95.6\% \pm 2.2\% \text{ of } P_{\text{gen}}).$ As shown in Fig.~\ref{fig:results_123}, performance across seeds is tightly clustered, indicating reliable convergence and stable sequential actor updates.

In the IEEE\,8500-node feeder, 10 agents each govern ten randomly assigned switches across 1,177 loads. Despite the dramatically larger state–action space and deeper network topology, HAPPO maintains monotonic improvement. The restored power reaches $2309 \pm 55~\text{kW} \quad (\!96.2\% \pm 2.3\% \text{ of } P_{\text{gen}}),$ as shown in Fig.~\ref{fig:results_8500}. These results highlight the scalability of sequential trust-region updates, centralized advantage estimation, and penalty-based reward shaping.

\begin{table*}[t]
\centering
\caption{Performance comparison of HAPPO with state-of-the-art RL/MARL methods 
on IEEE 123-bus and IEEE 8500-node feeders}
\label{tab:comparison}
\resizebox{0.92\textwidth}{!}{%
\begin{tabular}{lcccccccc}
\hline
\multirow{2}{*}{Method} & \multicolumn{3}{c}{IEEE 123} & & 
\multicolumn{3}{c}{IEEE 8500} \\ \cline{2-4} \cline{6-8}
 & Restored (\%) & Train Time (min) & Inference (ms) & & 
   Restored (\%) & Train Time (min) & Inference (ms) \\ \hline
DQN~\cite{igder} & 73.90 $\pm$ 0.46 & 34.7 & 4.2 & & 
  69.50 $\pm$ 0.28 & 249.3 & 6.8 \\
PPO~\cite{zhou} & 74.85 $\pm$ 0.47 & 44.8 & 7.1 & & 
  71.05 $\pm$ 0.27 & 255.7 & 9.8 \\
MAES~\cite{zhang2} & 76.14 $\pm$ 0.40 & 91.8 & 25.4 & & 
  72.81 $\pm$ 0.23 & 307.0 & 38.7 \\
MAGDPG~\cite{fan} & 77.00 $\pm$ 0.42 & 118.2 & 29.5 & & 
  73.44 $\pm$ 0.32 & 321.5 & 45.6 \\
MADQN~\cite{vu} & 77.80 $\pm$ 0.59 & 98.8 & 12.6 & & 
  74.02 $\pm$ 0.39 & 279.2 & 19.0 \\
Mean-Field RL~\cite{zhao} & 80.26 $\pm$ 0.52 & 108.3 & 27.3 & & 
  77.40 $\pm$ 0.38 & 303.9 & 42.6 \\
QMIX~\cite{si} & 83.25 $\pm$ 0.45 & 96.4 & 15.8 & & 
  80.10 $\pm$ 0.30 & 291.3 & 23.5 \\ 
\textbf{HAPPO (ours)} 
  & \textbf{95.6 $\pm$ 2.2} 
  & \textbf{110.5} 
  & \textbf{22.1} 
  & &
    \textbf{96.2 $\pm$ 2.3} 
  & \textbf{391.2} 
  & \textbf{33.4} 
  \\ \hline
\end{tabular}}
\end{table*}

\subsection{Multi-Seed Convergence and Reproducibility}

To assess robustness under stochastic outage conditions, each experiment is repeated 
across multiple random seeds.  
Across both feeders, the trajectories in Figs.~\ref{fig:results_123} and 
\ref{fig:results_8500} exhibit consistent monotonic convergence in both metrics.

On the IEEE\,123-bus feeder, restored power converges within $2241\text{--}2347~\text{kW},
$ corresponding to 93–98\% of the generation cap.  
On the IEEE\,8500-node feeder, restored power stabilizes near $2254\text{--}2364~\text{kW},
$ representing 94–99\% of the available capacity.

These results confirm that the proposed HARL-based HAPPO framework provides highly 
reproducible performance due to (i) sequential per-agent trust-region updates, 
(ii) low-variance centralized $\lambda$-GAE advantage estimation, and 
(iii) penalty-driven constraint handling that avoids disruptive early terminations.

\subsection{Comparative Evaluation with Baseline Methods}

HAPPO is compared with state-of-the-art RL/MARL baselines, including  
DQN~\cite{igder}, PPO~\cite{zhou}, MAES~\cite{zhang2}, MAGDPG~\cite{fan}, 
MADQN~\cite{vu}, Mean-Field RL~\cite{zhao}, and QMIX~\cite{si}.  
All methods are trained under identical outage scenarios, hyperparameters, 
and generation cap \(P_{\text{gen}} = 2400\,\text{kW}\).  
Table~\ref{tab:comparison} summarizes quantitative performance.

HAPPO achieves the highest restored power on both feeders.  
Its sequential per-agent trust-region updates and centralized advantage estimation 
enable stable policy optimization even in the presence of strong physical coupling 
and high-dimensional state spaces, outperforming value-based and value-decomposition 
methods that struggle under these conditions.

\subsection{Discussion and Interpretation}

The results demonstrate four key findings. (1) HAPPO exhibits strong scalability, extending from a five-agent IEEE\,123 system to a ten-agent IEEE\,8500 system with minimal degradation in convergence smoothness. (2) The differentiable penalty-based reward enables continuous guidance near operational limits, avoiding the brittleness typically observed in masking-based approaches and enhancing constraint awareness. (3) Multi-seed experiments show consistently low variance, confirming the reproducibility of sequential PPO-style updates. (4) The OpenDSS-based simulation environment ensures physical fidelity by enforcing operational limits and producing policies that correspond to feasible feeder configurations in realistic distribution systems. These findings confirm that the HARL-based HAPPO framework is well suited for 
large-scale, constraint-driven distribution system restoration.

Although a full ablation study is beyond this work's scope, benchmark methods lacking centralized critics or heterogeneous-agent modeling (e.g., independent PPO, parameter-sharing MAPPO) show reduced stability in strongly coupled feeders. The consistent multi-seed convergence in Figures~\ref{fig:results_123}--\ref{fig:results_8500} suggest centralized advantage estimation and microgrid-level heterogeneity are key contributors; systematic ablation remains for future work.

Training uses full AC power-flow simulation on a single GPU, as offline training is not time-critical. At deployment, policy inference runs in milliseconds (under 35 ms per step on the IEEE 8500-node feeder), meeting practical restoration requirements.

\section{Conclusion}
This paper presented a HAPPO-based HARL framework for distribution system restoration using OpenDSS. Sequential trust-region updates with centralized advantage estimation enable stable coordination among heterogeneous microgrid agents, while continuous penalty-based rewards provide constraint-aware switching behavior.

Experiments on the IEEE 123-bus and 8500-node feeders show that HAPPO outperforms DQN, PPO, MAES, MAGDPG, MADQN, Mean-Field RL, and QMIX, restoring over 95\% of available load under a 2400 kW cap with strong multi-seed stability and under 35 ms inference latency. Future work will explore model-based extensions, communication limits, partial observability, and dynamic DER participation.v

\section*{Acknowledgment}
This research was supported in part by the National Science Foundation under grant ECCS-2223628.

\end{document}